# Tangles: Unpacking Extended Collision Experiences with Soma Trajectories

STEVE BENFORD

School of Computer Science, The University of Nottingham

RACHAEL GARRETT

Media Technology and Interaction Design, KTH Royal Institute of Technology

CHRISTINE LI

School of Computer Science, The University of Nottingham

PAUL TENNENT

School of Computer Science, The University of Nottingham

CLAUDIA NÚÑEZ-PACHECO

Faculty of Technology, Malmö University

AYSE KUCUKYILMAZ

School of Computer Science, The University of Nottingham

VASILIKI TSAKNAKI

Digital Design Department, IT University of Copenhagen

KRISTINA HÖÖK

Media Technology and Interaction Design, KTH Royal Institute of Technology

PRAMINDA CALEB-SOLLY

School of Computer Science, The University of Nottingham

JOE MARSHALL

School of Computer Science, The University of Nottingham

EIKE SCHNEIDERS

School of Electronics and Computer Science, The University of Southampton

KRISTINA POPOVA

Media Technology and Interaction Design, KTH Royal Institute of Technology

JUDE AFANA

School of Computer Science, The University of Nottingham

We reappraise the idea of colliding with robots, moving from a position that tries to avoid or mitigate collisions to one that considers them an important facet of human interaction. We report on a soma design workshop that explored how our bodies could collide with telepresence robots, mobility aids, and a quadruped robot. Based on our findings, we employed soma trajectories to analyse collisions as extended experiences that negotiate key transitions of consent, preparation, launch, contact, ripple, sting, untangle, debris and reflect. We then employed these ideas to analyse two collision experiences, an accidental collision between a person and a drone, and the deliberate design of a robot to play with cats, revealing how real-world collisions involve the complex and ongoing entanglement of soma trajectories. We discuss how viewing collisions as entangled trajectories, or 'tangles', can be used analytically, as a design approach, and as a lens to broach ethical complexity.

## 1 INTRODUCTION

Collisions in Human-Robot Interaction are overwhelmingly considered something to be actively avoided or, in their eventuality, quickly mitigated. Such strategies may be appropriate for the applications that have driven robotics developments to date such as industrial robots in factories and autonomous vehicles on roads. However, we argue that they will be insufficient to deal with the ways in which humans might encounter physically actuated technologies as they spread into more chaotic and uncontrolled environments. Robots are moving from factories into homes; autonomous vehicles are migrating from cars to wheelchairs; children and pets are engaging with robot toys. The spread of ever more mobile and autonomous technologies into our domestic lives brings to the fore the matter of *how,* not if, people will collide with them. In such environments collisions are typically a mundane and routine affair, sometimes a necessary one as when jostling past or wrangling objects, and sometimes even a desirable and joyful one as in children's play and adult sports.

  We therefore undertake a radical reappraisal of how humans collide with robots and similarly actuated interactive technologies: one that reconsiders and even embraces collision as a vital, necessary facet of human bodily experience and consequently establishes it as a powerful resource for design. By collision, we mean relatively forceful, sometimes even violent, contact between humans and robots. Collision is different from other kinds of touch in its level of force; potential consequences for the parties involved such as the possibility of pain and even injury; and somewhat ballistic, unpredictable and uncontrollable nature. Collisions may be accidental or planned. They may be quickly forgotten or remembered for a long time. From an HCI point of view, they can be extremely powerful user experiences. Unsurprisingly however, given their perceived undesirability, beyond risk assessment, there are few tools or approaches for designers to engage with, analyse, or design for collision experiences [39].

Our goal is to provide HCI and HRI researchers with resources to broach the design space of collision. The authors of this paper are an experienced group of HCI/HRI researchers and designers, with an interest in somaesthetics [24] and uncomfortable interactions [6]. We take as our point of departure that collisions are inherently somaesthetic experiences; actively involving one's body-mind; encompassing a spectrum of fun, thrilling, or discomforting experiences; but also in need of careful attentiveness. First, we turned to soma design methods to attend to the somatic experience of colliding with robots. We conducted a two-day soma design workshop where we sensitised ourselves to different qualities of collision with the help of expert practitioners [55]. We then undertook a technological exploration with a series of autonomous and robotic technologies—telepresence robots, mobility aids, and a quadruped robot—to ideate different ways of colliding with robotic bodies. Following the workshop, we employed the established concept of soma trajectories [60] to represent our collision experiences, identifying key transitions between consent, preparation, launch, contact, ripple, sting, untangle, debris and reflect. We then decided to use these collision trajectories to analyse data from two "real" collision experiences, i.e., interactions that were not heavily orchestrated in the context of our workshop. These two case studies—an accidental collision between a person and a drone [46] and intentional collisions arising from cats playing with a robotic arm [56]—revealed additional layers of complexity to collision experiences, leading us to consider how collisions involve complex entanglements of soma trajectories. Such *tangles* encompass the nine key stages of the collision trajectory as well as the notions of entanglement, disentanglement and re-entanglement to highlight the ongoing physical and emotional relationships between various parties. Finally, we reflect on how considering collisions as tangles can help (i) analyse collision experiences to unpack the qualities of desirable or undesirable collisions and how to enable or recover from them; (ii) generate potential design approaches for collision experiences; and (iii) offer a nuanced lens to understand the ethical complexities of colliding in different contexts involving different stakeholders, consent and withdrawal, as well as care and repair work.

## 2 RELATED WORK

We consider related work in two parts. First, we cover literature within robotics, HCI, and related fields that largely views collision as a problem to be avoided or mitigated. We then turn to a body of literature from beyond these technical fields that motivates why we should reconsider collisions as a complex topic of import within HRI and as an important and potentially beneficial facet of human experience.

### 2.1 Avoiding and mitigating collisions

In robotics, the dominant focus has been on avoiding collisions between robots and humans. This is understandable when industrial robots are placed near human operators. Lasota et al. summarize this well in their extensive survey of work in Human-Robot Interaction (HRI):

> "In simple terms, in order for HRI to be safe, no unintentional or unwanted contact can occur between the human and robot. Furthermore, if physical contact is required for a given task (or strict prevention of physical contact is neither possible nor practical) the forces exerted upon the human must remain below thresholds for physical discomfort or injury. We define this form of safety in HRI as physical safety." [35]

Ensuring such safety is done in several ways – creating hard and soft fences for the human and the robot, so that they don't occupy the same space; defining safety zones for the human and limiting the velocity and force of the robot in case of safety hazards to allow time for relocation; avoiding collisions through detection of obstacles, including nearby humans; motion planning to avoid objects and humans, and multi-task optimisation, including prediction of human motion. There are also 'post-collision' safety methods, in which the robot changes behaviour once it detects that a collision has occurred [35]. If the robot is small enough, for example a cleaning robot, low impact collisions with objects and humans become a strategy for detecting then moving away from them.

The spread of autonomous mobile delivery robots, telepresence robots and other mobile robots into crowded and 'messy' human spaces, makes collisions with humans inevitable. In these contexts, research into autonomous robot navigation has turned to the question of social as well as physical acceptability. Not only should robots avoid obstacles, but they should consider the comfort, naturalness, and sociability of their manoeuvring [34], treating people as social beings and not as mere obstacles [31], for example by recognising and respecting their personal space [49]. Proposed solutions include algorithms based on the Social Force Model [23, 31], that balances attractive forces exerted by navigation goal points with repulsive forces emanating from nearby humans. However, this can result in indecisive collision adverse robots becoming stuck in local minima positions, especially in crowded spaces [59]. Autonomous systems may also exhibit less desirable social behaviours such as hesitancy, dithering, and poor social signalling. Brown's study of partially self-driving Tesla and Google cars reported how their inability to interpret and convey the subtle social cues that human drivers employ when deliberately leaving or closing gaps for others led to confusion [8].

Other emerging applications of robots require collaboration between robots and human operators, for example assistive robots caring for people in their homes helping them get dressed [14]. In this situation, contact between robots and humans is necessary and inevitable; however, the focus remains primarily on keeping this contact 'safe'; i.e., on limiting force involved in contact and avoiding any collisions with humans [42] rather than on considering other important facets of human touch. Rode at al's study of children navigating a robot through a crowded obstacle course showed how the inevitable collisions could be difficult for the robot operators to disentangle, requiring physical intervention from human wranglers [50]. The desire to collaborate with robots to deliver forceful actions such as drilling, sawing and cutting has motivated techniques for planning human manipulation to minimise muscular effort [18] and adapt to varying external forces [11].

Finally, we note that collisions need not involve direct physical contact to be consequential. Sanders showed how near misses influenced cyclists' perceptions of risk, leading to reduced cycling [53]. Trump and Parkin's study of road traffic incidents involving horses revealed that 44.5 percent of injuries resulted from incidents with no impact with the horse, confirming that near miss and post-avoidance collisions can still have significant physical consequence [62]. In a different vein, studies of collisions in virtual reality in which there is no physical contact have revealed how 'collision anxiety' can still adversely affect user experience and erode confidence in the technology [48].

In summary, humans are increasingly likely to experience collision with a raft of interactive technologies that are finding their way into our world. Even if one views such collisions as problematic, it is important to understand and design for the specific aesthetic qualities of impact and so maximise the potential to swiftly recover from them, minimise danger and significant discomfort during the aftermath, and learn from them. Attending to the experience of accidental collisions and their aftermath can therefore be viewed as designing for breakdown in the best way possible. However, there is a body of work beyond robotics and HCI that points to potential benefits of collision as an experience, and it is to this that we now turn to motivate the case for embracing collisions in some interactions and contexts with technologies, not purely avoiding and mitigating them.

**2.2 Collision as a necessary and positive facet of human experience**

As somatic creatures living in a physical world, collisions may be requisite to our growth and wellbeing. This is well-established in the body of literature on childhood development. Multiple authors [9, 54, 44] describe the importance of risky play in childhood. Such play can involve risk of physical injury from heights, speed, and objects. Playground games often involve forceful physical collisions, such as knocking people down or grappling them. Such play is inextricably linked to collisions, from falls or injuries from the play environment, to touching and hitting in chasing games such as Tag, to full body collision and tackling in games such as British Bulldog [51]. Though individuals have different tolerance levels, risky play is universally sought after and has been shown to help not only develop physical fitness, but also emotional and social skills such as self-regulation, empathy and problem solving [9, 44]. Conversely, it has been suggested that the historical rise in safety precautions preventing this type of play may have contributed to the endemic rise in mental health problems such as anxiety, depression, poor emotional regulation, over-dependence, and feelings of helplessness [21]. It has also changed the societal view of children from being relatively capable to incapable and in constant need of protection. The increasingly ubiquitous practice of 'soft play' has gone some way to reestablish collision as a fundamental aspect of play. Soft play creates environments where it is comparatively safe and actively encouraged for children to collide. However, the same treatment is not yet given to technology, where avoidance remains the norm. Given such arguments, we speculate that the ubiquitous deployment of collision avoidance technologies might constrain peoples' somatic abilities to understand the world around them and their place in it. We further speculate that embracing physical play with robots could potentially enhance soft play, for example in making it adaptive and inclusive for children with different capabilities.

For many, the desire to collide persists into adult life, most notably in sport. Interestingly, participation in such activities is not shown to be correlated to levels of aggression [30, 32], suggesting that collision may be an enjoyable thing in and of itself. Participants obviously have a risk of causing or receiving pain and suffering, however, work on acute and chronic pain shows that pain is not merely a physical phenomenon resulting from bodily harm, but a somatic experience significantly influenced by thoughts, emotional state, and context [27, 28, 41]. Furthermore, there is evidence pointing to extensive overlap in neurological processes and structures between pain and pleasure [36]. In cases such as sport, pain can be rewarding both at the time, and in retrospective enjoyment of having overcome it [7]. The example of sport also highlights the risks of hastily averted collisions, pulling out of a collision – for example a tackle – can be more dangerous than following through. Literature in sports psychology similarly suggests that participation in extreme sports, for example, is not indicative of careless or risk-taking personality traits [17, 11]. Rather, such participants tend to be meticulously prepared, highly aware of their physical and mental limits, and view their participation in such activities as connected to a search for human values: freedom from constraints, freedom as movement, freedom as letting go of the need for control, freedom as the release of fear, freedom as being at one with the natural environment, and freedom as a matter of choice and personal responsibility [11]. The attitude and preparedness of such athletes suggests that safety and risk do not necessarily exist on opposite ends of a spectrum.

These studies suggest that collision is a more complex phenomenon than is generally considered in HRI, entwined with complex ethical discussions about our freedom to learn about our bodies, express ourselves somatically, and to act freely within the world. We stress that we are not making an argument for unlimited freedom as there are boundaries to what is morally right within the context of both society and design practice. However, this literature suggests that we as designers need tools – ways to engage with, analyse and discuss the experience and ethics of collision – to properly broach these complexities to consider *if or when to enable or restrict collision through design*. For example, playful colliding, especially that done by children as discussed above, is an aspect of their right to learn and grow in the world [17]. Although children should clearly be protected from situations where they are at risk of becoming seriously hurt, care must be taken not to harmfully restrict their right to play. Play can be seen as a child's freedom and a fundamental part of their development [22]. Adults should also have the right to play – no matter whether rich or poor [17]. Previous HCI research has recognised that many adults exercise their right to deliberately engage in uncomfortable experiences, such watching scary films, participating in provocative performances and installations, riding rollercoasters and even engaging in religious rituals in the interests of entertainment, enlightenment or social bonding, inspiring the design strategy of 'uncomfortable interactions' [6].

In summary, we argue that most of the research on collisions in HCI and HRI has been driven by the development of autonomous vehicles and industrial robots. This has understandably resulted in collisions and similar physical contact with technologies being treated as something to be wholly avoided, which has already resulted in concerns about how people and autonomous vehicles share public space [45, 1, 12]. However, with the growing ubiquity of smaller robots and autonomous systems in domestic contexts, designers need to reassess the complete avoidance of collision. Hence, there is a need for tools to broach this complex space. On the one hand, there remains a need to anticipate, avoid and mitigate potentially dangerous collisions – considering multiple physical and contextual factors such as velocities, differences in mass, compliance of colliding surfaces, textures and other materials factors to name but a few. On the other hand, there is a need to consider the design of everyday collisions that enable us to learn about our bodies and world, experience pleasure and joy, and maintain our freedom to engage with the world. This leads us to argue for a paradigm shift in designing collisions, one that embraces collision as an important aspect of engaging with robots and that balances benefits and risks in a manner that, as Brussoni et al. write, keeps people 'as safe as necessary' rather than 'as safe as possible' [9].

## 3 METHOD OVERVIEW

We have chosen to present our approach in four stages. First, we provide an overview of our initial workshop, detailing our somatic and technological explorations with a series of robots and autonomous systems. Second, we show our application of soma trajectories to identify the nine key transitions of a collision experience. Third, we present the analysis of two case studies using these transitions. Finally, we discuss the emergence of the notion of *tangles*. It is perhaps a little unusual to present the generation of concepts in this iterative way. However, we believe it to be useful to 'show our workings', detailing how this concept iteratively and gradually evolved over time, gradually emerging from our design work and analysis through a series of discussions around an online (shared) whiteboard as we sketched and refined emerging concepts. We see this as characteristic of authentic reporting on the research through design processes, where messy practice-based explorations are iteratively refined into more theoretical knowledge [33].

The **first** stage (*Section 3 – A Soma Workshop to Explore Embodied Collision*) involved engaging with soma design methods. S*oma design* is a stance that involves using the lived, corporeal body as a resource for design work, facilitating encounters between materials and bodies to reveal a deep understanding of the aesthetic relationships between the two [24]. Theoretically, soma design strives for a deep and holistic appreciation of how the mind-body experiences interaction, which we feel to be highly appropriate for understanding the nuances of collision as a human experience. Intensive soma design workshops have been shown to be a rich way for researchers to explore concepts together from a somaesthetic perspective [40, 26, 61]. Practically, soma design may combine various methods, typically involving gathering a group of experts both in the relevant domain and in soma design, often beginning with some orientation or sensitising activities, then conducting material and technological explorations [63] usually using rapidly made low fidelity prototypes, wizard of Oz techniques [13], or tangible toolkits like soma bits [64]. Each of the activities are documented and articulated using a range of techniques including group discussion, body maps [37] and soma trajectories [60]. After the workshop, participants reflect on the experiences and findings, often as a group (e.g. in [61]), writing detailed accounts and drawing out key findings.

The **second** stage (*Section 4 – From Contact to Collision Trajectories*) is the starting point for *tangles;* the recognition that collisions should be treated as extended and gradually unfolding user journeys. We turn to an existing conceptual framework called *interactional trajectories* which expresses how interactive user journeys unfold over extended time periods [5] as a foundation upon which to build. Interactional trajectories were initially introduced to express the complexities of mixed reality performances [4] but have since been applied to a variety of areas within HCI, including to somaesthetic design through the idea of soma trajectories [60]. Key to our work here is the idea that interactional trajectories negotiate various transitions along the way, significant moments in the user journey that need to be carefully designed or managed. They are also socially interleaved to represent how various participants' journeys converge and diverge over time. By reflecting on our soma workshops, we derive nine transitions to help us better understand collisions as unfolding trajectories.

The **third** stage (*Sections 5 and 6 – Collision Case Studies*) involved applying these collision trajectories and transitions to analyse two examples of "real-world" collisions – collisions that were not heavily orchestrated in the context of our workshop – but rather collisions about which we had existing data from other research activities. The first involved the accidental collision between a person and a drone that had been caught on camera during a recorded design activity [46]. The second involved the deliberate design of a robot to play with cats that had been created as part of an interactive art exhibit by the artists Blast Theory [57].

The **fourth** and final stage (*Section 8 – Tangles*) involved reflecting on these analyses to further unpack the complexities of collisions, considering how they involve ongoing entanglement, disengagement and re-entanglement; how these span physical and emotional dimensions of experience; and how they can draw in multiple parties beyond those immediately making contact.

We present our process in these four distinct stages for the purposes of clear communication and readability. However, our explorations of collisions iterated between soma design workshops, reflections on these, and testing emerging concepts by applying them to the analysis of other studies of experiencing and designing collisions with robots.

## 4 A SOMA WORKSHOP TO EXPLORE EMBODIED COLLISION

In this section, we present an overview of our workshop activities and describe the technological explorations with three robots/autonomous systems: telepresence robots, mobility aids, and a quadruped robot. Each day of our two-day soma workshop began with a movement expert facilitating sensitising exercises to help orient us to the experience of collision. On the first morning we were joined by a Feldenkrais practitioner. Feldenkrais is an activity commonly used in soma design [25] as it helps participants focus inwards, considering the connectedness of their bodies in detail and exploring how movements feel. Their first exercise involved lying on the floor and allowing our extended legs to drop, folding at the knee to collide with our thighs. Repeating this simple collision enabled us to tune into important facets of the collision experience: the discomfort of the hold, the anticipation of the drop, the moment of release, the point of no return, the impact, the spring effect of the knee and the ripple effect on the leg. Next, we performed an exercise where a wooden rod was placed between the index fingertips of two participants with the goal of moving around while keeping the rod in unstable equilibrium between them – sometimes even with eyes closed. This sensitised us to aspects of shared collision experiences including leading and following, movement negotiation, and collision avoidance or near missing (trying not to drop the rod).

On the second morning, we were joined by a professional choreographer and dance teacher. He first invited us to explore different types of blows (stinging, glancing, and direct) and to feel the 'echo' of a collision by striking our own exposed skin in different ways. Here we noted, with surprise, how long the sensory echo of slapping the skin of our arm would last. We also noted the experiential differences between a hit on the *skin* versus hitting across a *muscle* versus a hit on a *bone* structure, sensitising us to the notion of the 'depth' of a collision. For example, blows might result in a sharp but shallow pain that spreads out across the surface of the skin but does not travel deeper; might reverberate through the muscle in a deep, resonant manner; or might be highly concentrated and pierce down to the bone. Next, he invited us to create a collaborative collision experience with a partner, before switching partners to share our experiences with others: examples of these included a lightly choreographed martial arts fight, a pat-o-cake children's game (involving choreographed rhythmic slapping of hands), and one partner very slowly moving their index finger towards their partner's face before surprisingly striking their nose. These experiences highlighted preparation, coordination, suspense, surprise, ambush and the vulnerabilities and sensitivities of different parts of the human body.

These sensitivities were carried forward into low-fi prototyping. We divided into several groups, each looking to design a collision experience around a different technology.

### 4.1 Telepresence Robots

The first group explored colliding with telepresence robots, which are remotely controllable mobile robots with Wi-Fi connectivity and video and audio capabilities, typically used for teleconferencing purposes. These robots were chosen as they are becoming increasingly ubiquitous in contexts such as crowded museums or event spaces. Under normal use, these robots are not intended to bump into objects, being equipped with sensors and algorithms to avoid obstacles and so allow inexperienced users to drive them safely. However, scenarios of telepresence robots colliding with humans are not uncommon. In contrast to this product-design intention, this group prototyped three experiences with a Double 3 telepresence robot from Double Robotics, which features a small two-wheel base above which a screen/camera is supported via a thin pole.

**Controlling and Being Controlled:** Inspired by the wooden rod exercise, we (the first group) employed bodystorming [43] to explore how to control the robot while slowly approaching collisions as a felt experience. One member of the team was blindfolded and became "the robot" surrendering their agency and allowing others to "drive"' them across the building and later outside. We attached vibratory soma bits [58] to both arms, which were activated by two other team members. A third stimulus on the back requested the human-robot to stop its march. This exercise reinforced the importance of anticipation, the sense of vulnerability and the protection of the body which is important to collision experiences.

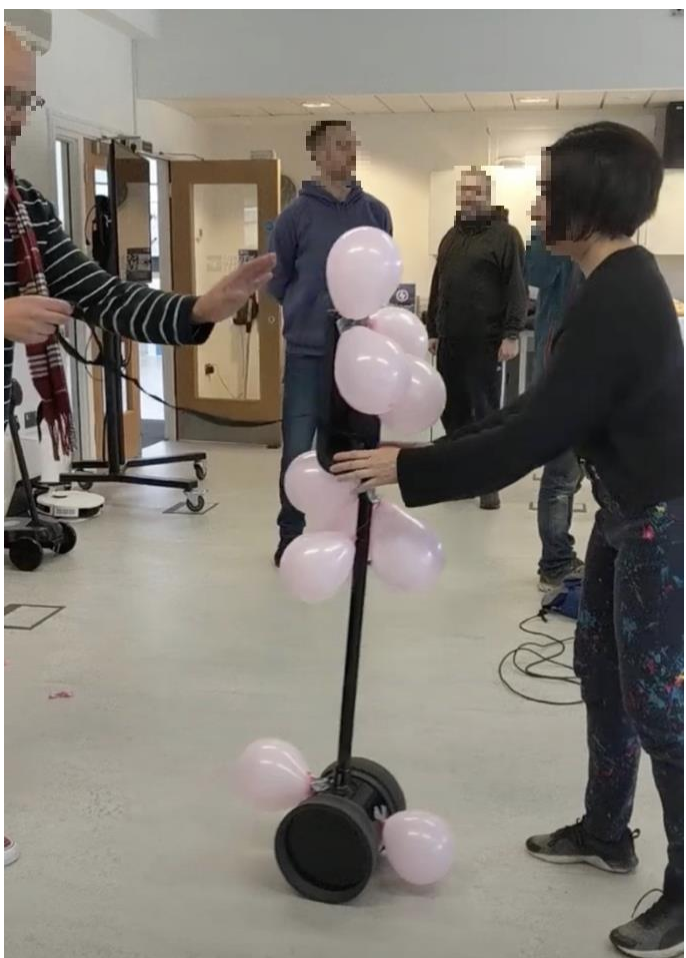

**Figure 1.** Exploring collisions using a leash and balloons attached to the Double telepresence robot

**The Leash:** Next, we wrapped a leash around the robot's "neck" to enable us to collide with it safely while minimising risk of damaging it (Figure 1). We soon discovered aesthetic qualities that made such collisions pleasurable and rewarding. The flat and smooth surface of the screen of Double 3 being pulled towards the torso resembled the gesture of a football player stopping the ball with the chest. Pulling the leash with determination towards oneself would cause the robot to accelerate hastily, enhancing the sense of anticipation while satisfactorily hitting the body with just "enough" pressure to feel the liveliness of its presence. Finally, the contact of the screen "being stopped" by the torso highlighted the differences between the organic presence (the body being bouncy and soft) versus the cold rigidity of the robot. After a collision, the robot would bounce back to be pulled and collide with the body once again, as if playing with a yo-yo (See Figure 2). However, the team member on the other side of the computer — who normally would control the telepresence robot — described being awkwardly exposed to the coming and going of body parts as captured by the robot's camera leading us to reflect on how we were not only limiting the affordances of the robot but also taking away the agency of the driver, who also needed to be considered in the collision experience.

**Shape-Shifting:** We then focused on the Double 3's body, ideating shape-shifting interactions where it would inflate and deflate, not only for safety purposes — and therefore protect the robot from the consequences of a collision — but also to indicate to others the social boundaries of the driver on the other side of the screen. We envisioned this inflatable body would also employ haptic receptors, involving the body of the driver more actively in the interaction and allowing the driver's intentionality to be expressed to some degree. Instead of colliding the robot against our bodies, we now used the leash to follow the directions of the driver. We prototyped this idea by attaching balloons to the robot (Figure 1), mimicked their dynamic inflation by using human touch on the driver to (i) hint about the proximity of other objects and (ii) suggest directionality to avoid collisions. After some testing, we felt that there was a clear gap between human touch (and the non-mandatory quality of subtly patting on the skin to attract some attention) and the very artificial felt qualities inspired by the robot. We came to the realisation that (gentle) human touch suggests rather than imposes, adding a layer of perceived agency during the interaction.

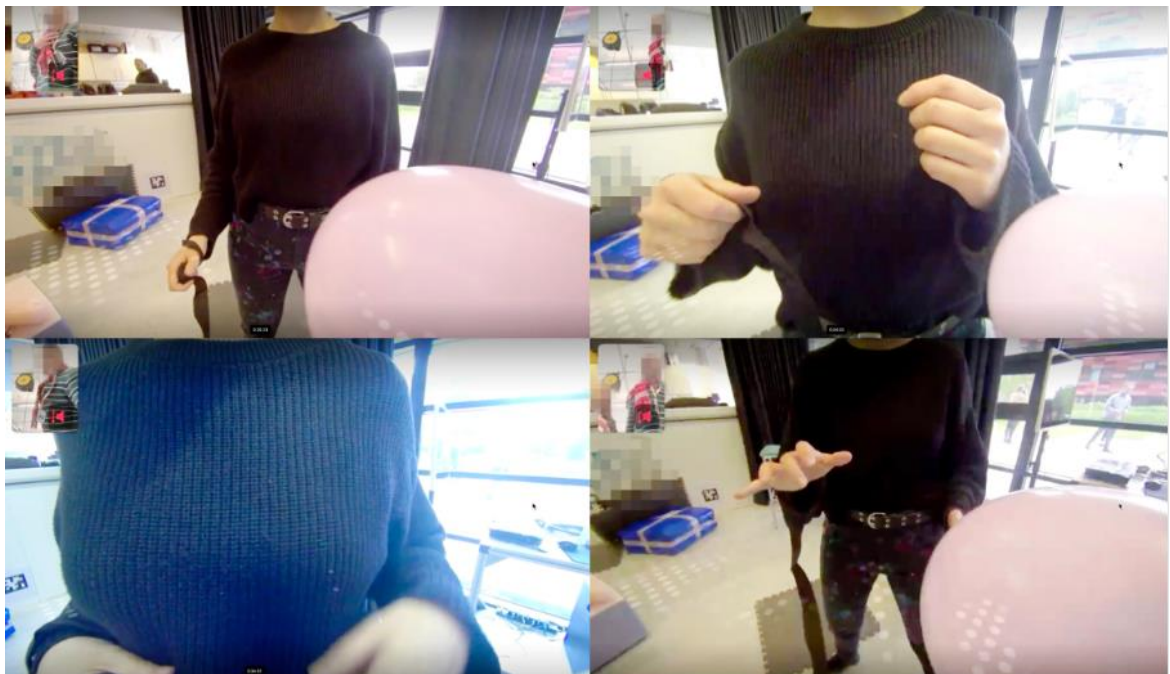

**Figure 2.** The leash around the robot resulted in a yo-yo effect, creating a fun experience for the one controlling the robot, but awkwardness for those on the other side of the computer.

## 4.2 Mobility Aids

The second group focused on robotic and non-robotic 'walkers' for older adults with mobility impairments. Though none of the researchers belong to this user group, one member conducts research in assistive technologies, so she helped the group to carefully navigate this design space. This technology was chosen as robotic mobility aids are increasingly common, not only for older adults, but also for children with disabilities and similar user groups. This group started with a presentation on the main issues users face when using mobility aids. The group then explored how collisions between mobility aids and the surrounding space (i.e., a wall, furniture, a bench) could be a way of enabling play while also extending one's perception of the physical world.

**Collision as a Playful Experience:** Inspired by the idea of enabling play between older adults and their grandchildren, we (the second group) improvised playful collisions between the walkers and furniture, then a ball moving on the floor. One person moved with the walker, while another maneuvered objects into close proximity ready for collision. We noted how curated collisions with an unfolding narrative – e.g., colliding with a surface that was softer, followed by progressively colliding with harder surfaces, or beginning colliding with smaller objects and progressively with larger ones – evoked feelings of playfulness and feelings of achieving closure, from anticipation of collision to the completion of the experience. We employed vibratory soma bits to highlight the sense of anticipation (see Figure 3) followed by a BHaptics Gaming TactSuit (which supports 40 localised vibrations to be varied in intensity and frequency) to imitate the texture of an object with which the mobility aid was colliding through vibrations felt on the lap, back, and feet, noting a strong feeling of connecting the body to the surrounding world through the mobility aid. Subsequent discussions raised the idea of a "touch-stroke-collide" approach to unfolding collision experiences. This entails how the gradual contact of the robotic vehicle with a surface could be mapped to a vibrational effect on the user's body through the haptic suit: encountering an object and slightly touching it through the proxy of the mobility aid would actuate a subtle vibrational effect, which would gradually increase its intensity felt on the wearer's body, as the object and the vehicle will collide with one another.

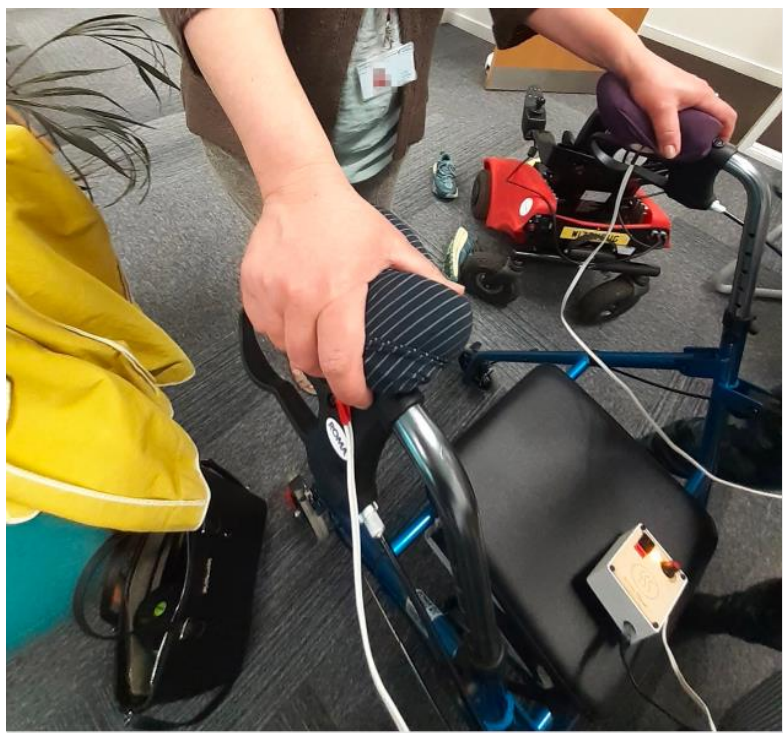

**Figure 3.** Using vibration on the handle of the mobility aid to transfer the anticipation of collision

**Collision Contributing to an Increased Somatic Literacy:** Inspired by how our Feldenkrais practitioner had drawn attention to micro-movements and our dancer had considered collisions as involving skin, muscle and bone, we extended our explorations of vibration to sequence and layer vibrations across the body. All participants in our group tried the TactSuit when walking with a mobility aid: one person was controlling the location of the vibrations and the person wearing the vest had their eyes closed as they moved towards and collided with objects (See Figure 4). The person controlling the vibration mimicked a ripple effect starting from the point of collision and slowly moving to another side of the body (horizontally, vertically, or diagonally). For example, when colliding with a sofa through the left wheel of the mobility aid, the ripple effect, or echo of the collision was transferred from the left leg to the torso, and then slowly to the right arm, which was experienced as a pattern of vibrations travelling diagonally across the body. We tried subtle and small types of collisions and intense or abrupt ones; varying the ripple effect of colliding depending on the type of material surface one collided with, the speed and intensity of the collision; and initiating the ripple effect from the part of the body nearest to where the collision happened and then spreading to the other direction on the body, and vice versa (creating sensations of sensory misalignment [61]. We reflected on how soma design speaks of helping people increase their somaesthetic awareness [24] considering how our approach might empower those with sensory or mobility impairments to reflect on their bodily sensations, actions and reactions.

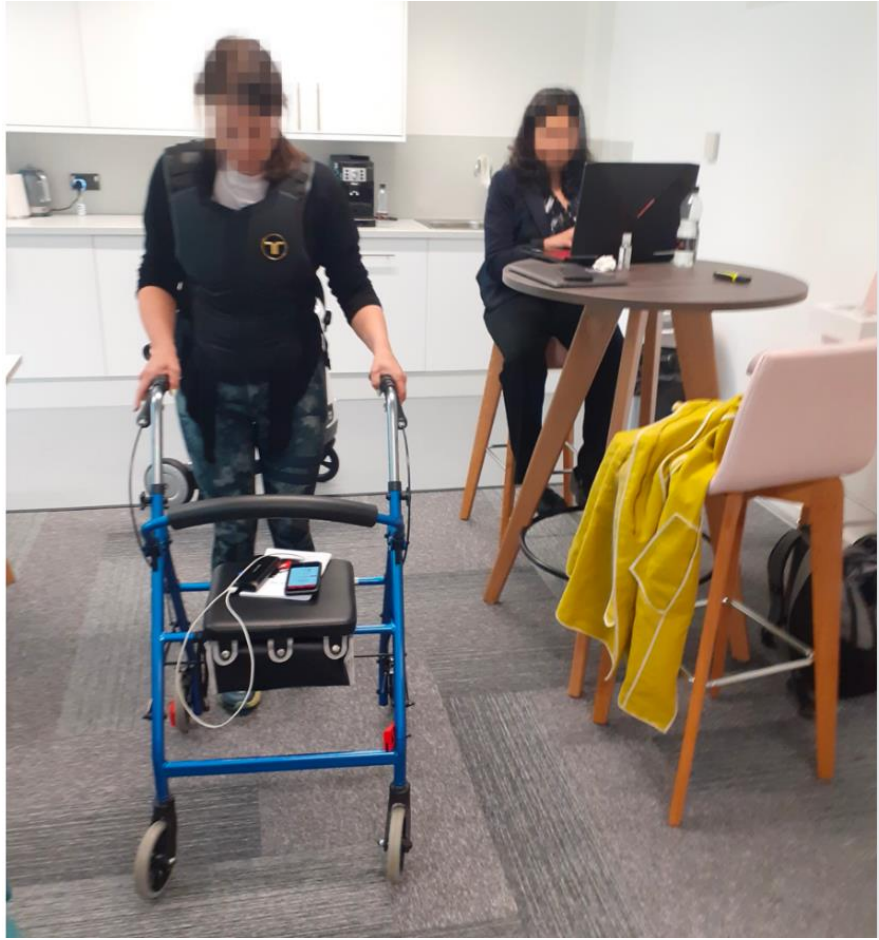

**Figure 4.** Mimicking a ripple effect of collision on the body through a vest with multiple vibrators. This ripple effect starts from the point of collision and slowly moves to another side of the body (horizontally, vertically, or diagonally)

### 4.3 Quadruped Robot

The third group explored collisions with the Boston Dynamics' quadruped robot, Spot, a dog-like robot which was originally developed to navigate challenging terrains and relay sensor data. Though Spot is not a domestic robot, a variety of zoomorphic robots designed as companions or toys are becoming more common. It was therefore decided to include the quadruped in these design explorations, which took place outside the studio in a small, paved courtyard surrounded by grass.

**Probing collisions with Spot:** Spot is a somewhat intimidating robot and so our initial explorations involved gentle probing of what it is like to collide with it. By default, Spot actively minimises or avoids running into collisions using its five cameras, which gives it a 360-degree field of view and a detection range of 4 meters. This is done by defining an "Obstacle Avoidance Cushion", which inflates the perceived obstacles and issues a repulsive velocity command to the robot to move away from collision. If Spot registers someone or something in its path, it will move either left or right to continue around the obstacle. However, this collision avoidance feature has its limitations and may miss objects that are very thin, short, elevated or transparent, and could result with unexpected behaviours when confronted with moving humans, as the default feature does not predict trajectories of moving objects. In addition, environment conditions, such as high contrast and inadequate lighting is expected to affect the collision avoidance performance unfavourably. We (the third group) began by exploring the limits of this feature, finding that long-grass and some low-objects confused the sensors. However, it as was difficult to orchestrate any meaningful collision with the robot actively programmed to avoid them, we elected to disable this feature. We then attempted to orchestrate a head on collision between Spot and members of our design team, with one team member controlling Spot through its live telepresence interface. One team member remained stationary, while the controller moved Spot towards them from different directions with direct and glancing angles until it collided. This drew our attention to the different characteristics of preparing for and responding to a collision that we felt the robot lacked: the quadruped did not alter its speed on the approach (speed up or slow down), did not alter its movement (try to dodge or attempt to hit), or respond to the collision (stop moving, retreat, acknowledge the impact). We also noted that while immediate collision was not painful, there was some risk of pain from secondary collisions where Spot might step on your feet while still trying to walk forward.

**Being Guided by Spot**: Inspired by the Feldenkrais practitioner's stick exercise, we tried an exercise where a member of our team was blindfolded and led around the environment by 'guide dog' Spot (remote controlled), while she maintained contact first via a solid stick and then a leash. Notable aspects of the collision experience were, on the one hand, being led into secondary obstacles by Spot, which felt like bumping into a table, but also led to a heightened sensory awareness of one's surroundings, and on the other, the possibility of getting tangled with Spot itself when close up, though it also felt good to lean into Spot when standing next to it as its body gently sways and then pushes back in an almost comforting way.

**Rough and Tumble Play:** Having built our confidence that we could collide with Spot, we set about exploring more playful experiences. We drew on the earlier somatic exercise involving different blows between skin, muscle and bone. We reflected that Spot's metal frame was mostly just 'bone' and decided to extend it with 'muscle' and 'skin'. We used the soma bits inflatables and regular balloons to create 'muscles' on its head and shoulders, both making them more inviting for collision but also allowing Spot to potentially signal its intent – perhaps inflating its muscles when in a playful mood. We used Velcro to add rough and sticky fur to Spot's sides, top and knees, and also attached Velcro to our own legs and lay objects such as balls so that things (including us) could stick to Spot, opening up more play opportunities (trying to attach and detach things) and showing where it had previously experienced collisions (like a dog's fur might collect bits of vegetation from a romp through the bushes). Finally, we experimented with several waggy and whippy tails—appendages attached to Spot at various points that would thrash around somewhat chaotically when he wagged his body, dramatically extending the possibility of colliding with him. For example, a tennis ball, on an elastic string, attached to a flexible rod mounted on Spot. We used this faster, more chaotic, unpredictable and yet softer kind of collision to play a game where the aim was duck beneath Spot's tail dodging close enough to its body to be able to stick-on or retrieve a Velcroed object. As when playing tennis, collisions with the ball were acceptable, if not always desirable, offering a playful balance of risk and reward. Moreover, Spot's remote operator was able to improvise a variety of movements that were perceived differently by observers from the gentle (moving slowly and smoothly) to the wild, aggressive and challenging (moving quickly and sharply).

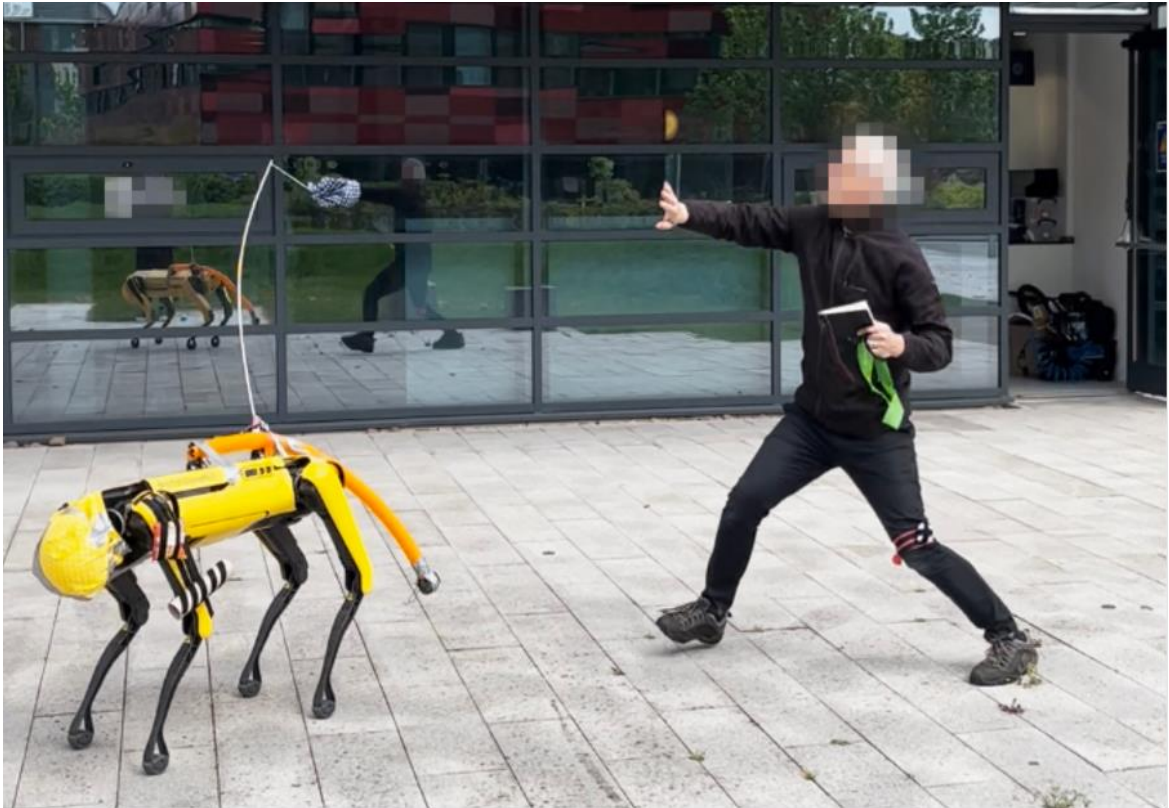
**Figure 5.** A 'whippy tail' with a soft weight on the end extended Spot's body and enabled more chaotic, playful collisions.

### 4.4 Workshop Reflections

These explorations sensitised us to the many complexities of collision as a rich phenomenon. We established that collision experiences begin long before any moment of physical contact and continue long after, with different stages of a collision having a profound effect on the experience. We sometimes undertook extensive preparations for collisions, including negotiating consent between the parties involved, while at other times collisions could be experienced as surprising accidents or even ambushes. We noted how contact may involve skin, muscle, bone and various attachments. We observed the many consequences of collision; how bodies ripple and sting, but that a valid collision experience might not involve direct contact at all; a near miss can deliver a powerful and consequential experience of colliding. We noted how collisions could involve multiple participants including local and remote humans, autonomous and teleoperated robots, and spectators who may ultimately become involved, and how different parties require disentangling, and how debris may lead to further collisions. Our explorations explored the different desirable and undesirable aspects of collision; for example, how a robot might move to encourage physical and even playful contact with these technologies, and how it might try to ensure these experiences were pleasurable and safe. From our somaesthetic perspective, we realised that collisions are experienced simultaneously in the mind and body; that we feel them emotionally as well as physically. Returning to our motivation to provide HRI with tools to systematically engage with, discuss, and analyse these complex collision experiences, we now turned to soma trajectories to represent these extended experiences.

## 5 FROM CONTACT TO COLLISION TRAJECTORIES

We reflected on the workshop over the course of several online meetings, trying to make sense of the various ways in which we had collided with robots. The core realisation to which we kept returning was that the overall experience of a collision extended far beyond the actual moment of physical content. Our initial bodily explorations had sensitised us to the lingering sting of being hit, to the emotional feeling of anticipating collision, had raised questions of consent, and even led us to try and design playful collision experiences for one another. Our subsequent experience with robots had further sensitised us the possibilities of extending both human and robot bodies to augment collision experiences and highlighted how collisions produce ripples and entanglements between parties and leave behind various kinds of debris. To bring coherence to these observations, we turned to the idea of 'interactional trajectories', a well-known framework for expressing extended user experiences involving technology, and in particular their extension into 'soma trajectories' which adopt a particular focus on bodily experience. Using an online shared whiteboard, we attempted to sketch generic examples soma trajectories to express the experience of collision, discussing potential labels for the different transitions along the way. In what follows we present the results of these discussions, using soma trajectories to visualise the unfolding of a collision in terms of nine key transitions: consent, preparation, launch, contact, ripple, sting, untangle, debris and reflect.

### 5.1 Collision as Soma Trajectories

As the starting point for analysing collisions with robots, Figure 6 presents an abstracted picture of a simple collision experience involving two parties (A and B) as being an interleaving of two soma trajectories. These are labelled with nine key transitions that we identified when reflecting on our soma workshop. The two parties might be a human and a robot, two humans (with technology mediating their collision), or possibly even two robots. We recognise from the outset (and shall see in the two case studies that follow) that this is a highly simplified picture in many respects. Complex collisions may unfold as a series of contacts involving ricochets and secondary collisions so that these lines may crossover several times. Transitions may occur in different orders, may overlap, may not appear at all (e.g., not all parties may consent to collision), or may be more ongoing (consent may be negotiated at various points). Nonetheless, we propose that this simplified picture provides a useful starting point for reasoning about extended collision experiences.

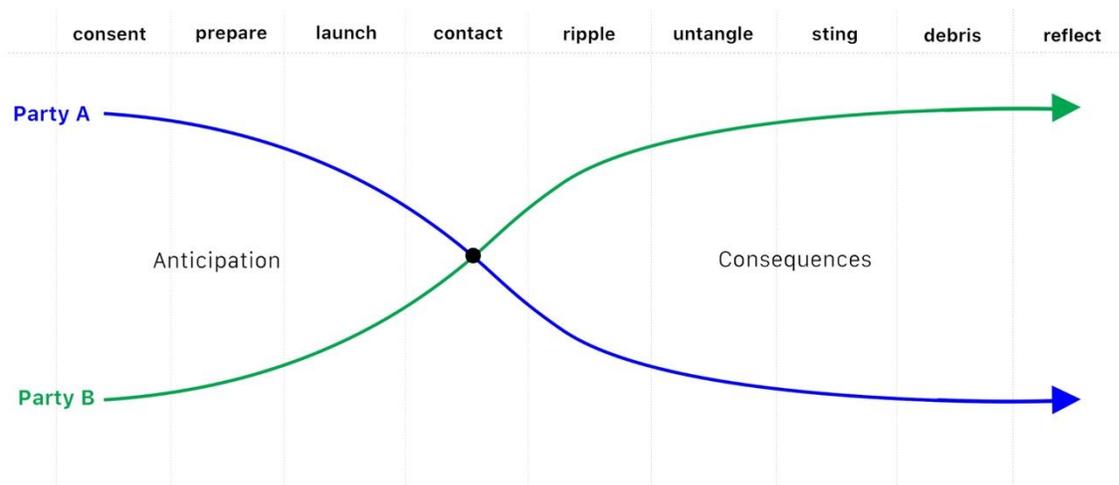

**Figure 6.** A generic and highly simplified collision trajectory

### 5.2 Collision Transitions

We identify nine key transitions along a collision trajectory, key moments that one should consider carefully when trying to understand or design a collision experience. While not every collision will overtly feature every transition (e.g., some may not involve consent or produce debris), we propose that the post-hoc analysis of collision experiences should consider to what extent they were present, while the design of future collisions and more collidable robots should consider to what extent each should be present. Three of these transitions (Consent, Prepare, Launch) contribute to the sense of **anticipation** prior to the moment of physical contact.

1. **Consent** *considers whether and how the parties involved agree to take part in the collision experience and how do they understand what it might potentially involve including acceptable boundaries.* Consent to collide may range from the explicit and formal such as when signing a waiver before undertaking a formally organised 'extreme' activity; to being written into organisational rules such as when taking to the sports field to play a governed and refereed contact sport; to being implicit such as when individuals who know and trust one another engage in playful rough and tumble. We recognise that collisions without consent may occur and might even be necessary, for example when urgently intervening to save someone from a worse accident. However, our focus is to encourage the recognition of negotiating consent as being an important and ongoing element of safe and rewarding collision experiences. While participants in our workshop had generally consented to be present and were involved in co-designing their own collision experiences, there were many specific moments when consent was clearly in play, such as when judging the boundaries of physical and digital engagement with various robots. Consent in these moments was sometimes verbally discussed but was often implicit, being signalled through adopting positions and postures that suggested an openness or otherwise to an unfolding experience.
2. **Prepare** *involves ensuring both physical and emotional readiness to collide.* Physical preparation may involve donning appropriate clothing, especially protective items, carefully checking and readying technologies, readying oneself by warming and loosening up the body, assuming an appropriate offensive or defensive posture, and choosing proximity to the approaching collision. Emotional preparations might include experiencing collision anxiety as reported in Ring et al's study of collisions in virtual reality [48] or even issuing threats and banter to exaggerate the consequences (e.g., wrestlers and boxers 'squaring off' in the ring). Preparation may also involve actively choreographing the upcoming collisions. Examples from our workshop include switching off collision avoidance modes on the Double 3 and Spot robots, attaching balloons and leashes to robots, and preparing spaces in which to collide.
3. **Launch** *reflects the idea that collisions are often ballistic to some extent, crossing a line beyond which they become unavoidable or inevitable.* Being unavoidable means that a party gives up the agency to choose to avoid the collision, entering a zone where it has passed beyond their control (perhaps they are too slow to react, for example). Being inevitable implies crossing a moment of physical commitment beyond which it is impossible (or at least riskier) to withdraw. Collisions that involve falling and leaping are prime examples, but other significant bodily actions such as punching, kicking and tackling also involve a sense of launch after which there is inevitable movement and follow through (hence the phrase to 'throw' a punch). There can also be a strong emotional aspect to passing beyond the moment of launch. Slow collisions involving considerable inertia (for example steering a boat into an obstacle) may involve a gradually dawning realisation of inevitability, while people sometimes describe time as slowing down during vehicle collisions. From our workshops, the Double 3, Spot, and mobility aids all have significant momentum and associated stopping distances, Spot's additional whippy tail can move very quickly and be impossible to duck, participants who were slingshot at the virtual wall experienced a point of unstoppable movement.

The ways in which these three transitions (Consent, Prepare, Launch) are present and interleaved will contribute to the overall anticipation of a collision, both in practical terms including ensuring safety, but also in experiential ones by enabling the impending experience to be potentially mitigated or, alternatively, savoured.

4. **Contact** *refers to the moment of direct physical engagement that occurs between the parties and how this is experienced by each.* This contact may be relatively violent in the sense that it exceeds gentle or normal touching, though this will inevitably be a subjective judgment. It may involve various parts of the body, some of which may be better equipped, naturally or artificially, to deal with them, while others may be more vulnerable. While it is generally a good strategy to protect vulnerable areas, certainly ones that can suffer injury, some may be sites for delivering intense collision experiences with minimal physical force and consequent risk (our improvisation exercises revealed even light tapping of the end of the nose to deliver a particularly intense experience). Moreover, not all collision experiences involve direct physical contact; near misses can be highly consequential. There were varied contacts to be considered in our workshops. The Double 3 exhibited a tendency to hit the physically and emotionally sensitive area of the chest. Spot's feet could trample on human feet, it was vitally important to avoid trapping fingers in its joints, and the whippy tail could cause a tennis ball to hit the head or torso. We also used vibration suits to try and enhance the sense of contact with the mobility aid for people who might not otherwise be able to sense them.

The potential violence of collisions that arises when two bodies try to forceable occupy the same space inevitably leads to **consequences** when the laws of physics confirm that they cannot. These can be expressed as a series of four further transitions (Ripple, Untangle, Sting, Debris) along the collision trajectory.

5. **Ripple** *considers how the impact of the collision subsequently plays out across the bodies involved and on the surrounding environments*. Limbs may flail and objects may fly, which may in turn lead to further ricochets so that a collision becomes a series of unfolding events. We note how ripples are often a necessary part of absorbing collision energy – without them, more serious damage may occur – as car designers will attest. However, the complex physical dynamics involved can be unpredictable, making it important to anticipate knock-on effects. Reflecting on our workshop, adopting a somaesthetic perspective on collision drew our attention to such ripples, from our preliminary exercises of raising and dropping our legs, through the process of trying out low-fi prototypes; the yo-yo rocking of the Double 3, Spot's body regaining its equilibrium causing his tail to flail around; and deliberately trying to invoke ripples on bodies using vibration suits or more simply hitting then with cushions.
6. **Untangle** *refers to both physically and socially disengaging from ongoing contact.* Untangling may be a cautious process where bodies have become tangled up and/or where parties need to cautiously assess whether any damage has occurred, and may involve assuming unusual postures, gradually withdrawing to normal personal space, removing protective items, and even withdrawing from the field of play. Rode at al's study of children navigating robots through an obstacle course revealed that significant work could be involved in untangling a collision [50]. Reflecting on our workshop, we sometimes needed to untangle the leashes on the Double 3 and Spot from other objects, untangle mobility aids from doors and furniture, or restore a semblance of order to our surroundings.
7. **Sting** *refers to residual sensations that persist after an impact, for example the mild tingling and burning of slapped skin that may persist for seconds or minutes afterwards.* There may also be an emotional sting from collisions in terms of the awkwardness, anger, hurt, guilt and shame experienced afterwards, or alternatively perhaps satisfaction in competitive sports. Stinging may also visibly reveal itself, for example through reddened skin. Beyond stinging, and generally to be avoided, lie scratches and bruises that may last longer and hurt more, and beyond these breakages that involve severe pain, long term trauma and possibly permanent damage. From our workshop, the effects of Spot stepping on one's toes or of being hit in the chest by the Double 3 could last for some time.
8. **Debris** *refers to the various objects, materials and other messes that may be left behind at the scene following a collision.* Debris includes items that become detached or ejected due to the collision which may require clearing up afterwards, which may in turn call attention to an otherwise unnoticed collision and/or involve some sort of consequential action from those involved. Debris might also provide souvenirs for post-hoc reflection, storytelling and accounting (historic trajectories), which may be important for both desired collisions (from reliving highlights to improving one's performance) and undesired ones (accident reconstruction as part of health and safety or legal processes). Debris from our workshops included balloons detached from The Double 3 and Spot's Velcro skin accruing objects.
9. **Reflect** *refers to thinking about a collision after the even*t. This can include official accounting for the causes and outcomes of the collision for the purposes of health and safety, insurance, and even legal proceedings. It can include more personal reflection in terms of apologies to others, telling 'war stories' of what happened, and analysing one's performance to learn how to collide better in the future (e.g. in sports). Finally, reflection is of course inherent to research, both into improving the technologies (e.g., developing new kinds of robots) and understanding human experience. Our workshops involved iterative reflection between tests in which we discussed what had occurred and devised further activities, followed by lengthy reflections over the course of months to inform the writing of this paper.

### 5.3 Socially interleaved collisions

A consideration of how multiple party's trajectories might interleave reveals various kinds of collision experiences as shown in Figure 7. **Mutually managed collisions** are ones in which both parties experience anticipation and consequences, passing through the above transitions, including mutually negotiating consent, careful preparation and extensive opportunities to assimilate what happened afterwards. Examples might be found in deliberately engineered collisions as part of sports and play. At the other extreme, **mutual accidents** are collision experiences in which neither party has much prior knowledge of the collision so that the experience begins with physical contact and tangling and unfolds from that point onwards as participants try and figure out and account for what happened. An **ambush** is a collision for which one party prepares while the other remains oblivious. A **near miss** is a collision in which both parties experience sufficient prior anticipation to be able to avoid physical contact, though there might still be ripples and other consequences afterwards. As we shall see below, collisions involving **teleoperation** like with the Double 3 robot are an especially interesting case as they might usefully be considered as involving three parties (the robot, its operator, and the colliding party) and hence three interleaved trajectories, such as the robot, its remote human operator, and other humans that it encounters physically. While we might want the operator's trajectory to closely follow that of the robot if they are to experience telepresence, in practice, there will be important ways in which their embodied experiences diverge. An **oblivious collision** is one where neither party ever becomes aware that they have collided, though there might still be consequences.

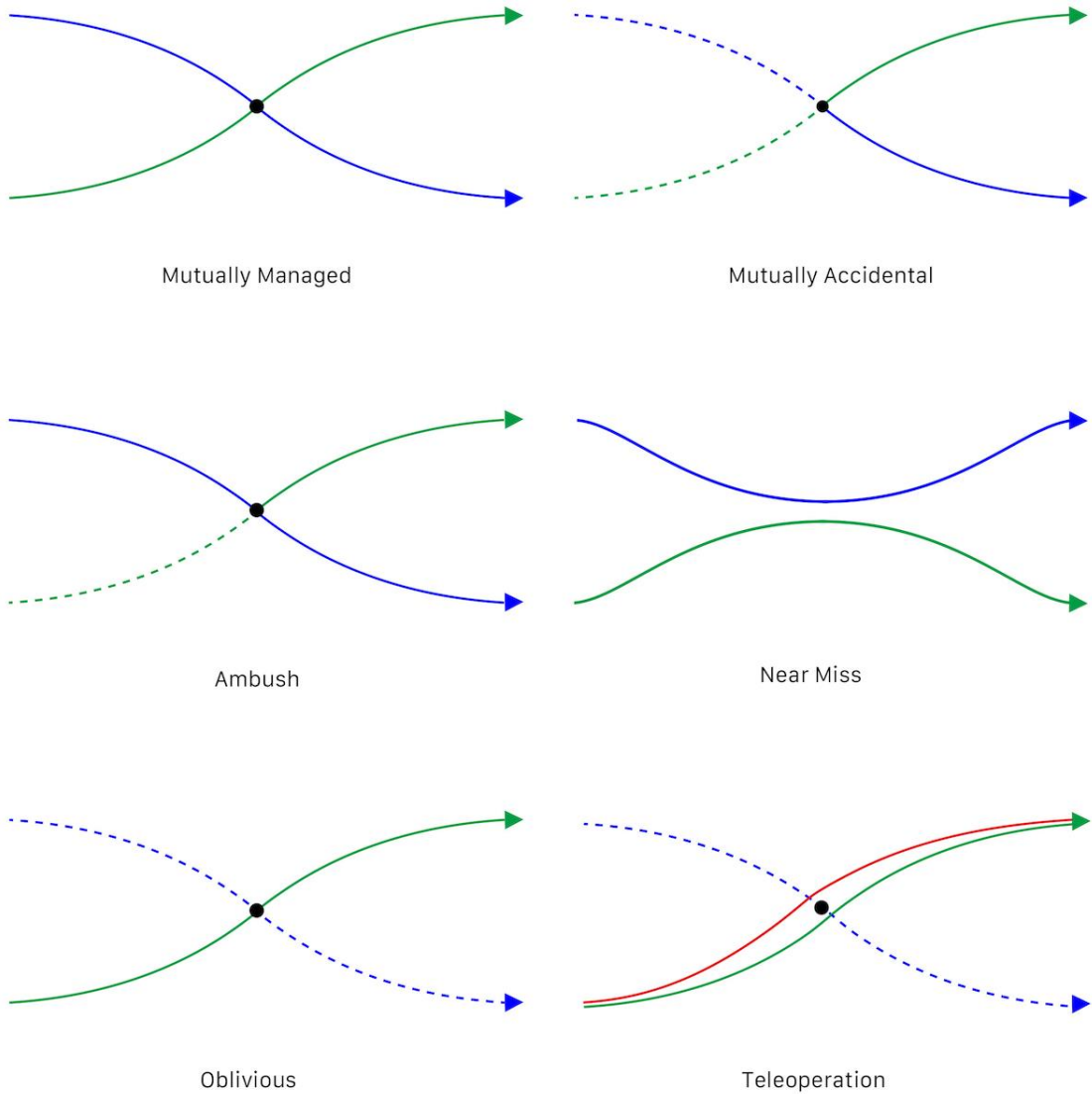

**Figure 7.** Mutuality in socially-interleaved collision trajectories

There are other significant roles, and hence other trajectories may become involved too. Collisions may be both orchestrated— controlled from behind the scenes—and spectated—observed by others who are not directly involved. Mutually managed collisions in sports provide particularly good examples of the importance of both roles, with referees managing safe collisions between players while onlookers vicariously appreciate them. Our workshop revealed the importance of various robot wrangler roles; for example, Spot's human wrangler or the local human companion who accompanied the telepresence robot. We suggest that such orchestration roles are likely to be present in many situations where robots are deployed, providing support for the various transitions and interleavings described above. They may help with negotiating consent and preparation, with warning oblivious participants, and with post-collision tidying up and accounting (including refereeing). Previous HCI research has distinguished witting spectators [47] who knowingly watch an event unfold from unwitting bystanders who may accidentally become involved [2, 58] Telepresence robots and mobility aids are likely to be deployed in the presence of both. As soon as we took Spot outdoors, we were approached by curious members of the public. Robots (and their orchestrators) need to be supported in involving spectators and bystanders in collision trajectories, for example helping them to understand what is going to happen or has happened while keeping them out of harm's way. We return to reconsider these additional parties to a collision when discussing tangles below.

Having established the basic shape of a soma trajectory for collisions based on our initial exploratory workshop, we next decided to text it against two 'real world' examples of collisions that we had experienced and documented in previous projects, returning to our shared whiteboard over the course of several further online meetings to try and systematically sketch and label detailed soma trajectories for these experiences.

## 6 COLLISION CASE STUDY: DRONE COLLISION

Here we present the first of two case studies using the trajectory and key transitions that we identified in the previous section. This case study (originally published in [46]) considers a mutually accidental collision that occurred during a design process exploring novel ways of interacting with drones. This incident was captured on a video recorder set up to record the design exploration. After reviewing the video footage, it was decided to conduct micro-phenomenological interviews with both Olivia[1] (the drone pilot) and Maria (the person into whom the drone crashed) concerning their respective experiences of the crash. This data was on hand for us to use as part of our analysis.

### 6.1 Description of Drone Collision

During a design exploration, Olivia, Maria, and Patrick bodystormed how one might control a drone using a shape-changing device that imitates the pressure being exerted on the drone as it moves through the air.

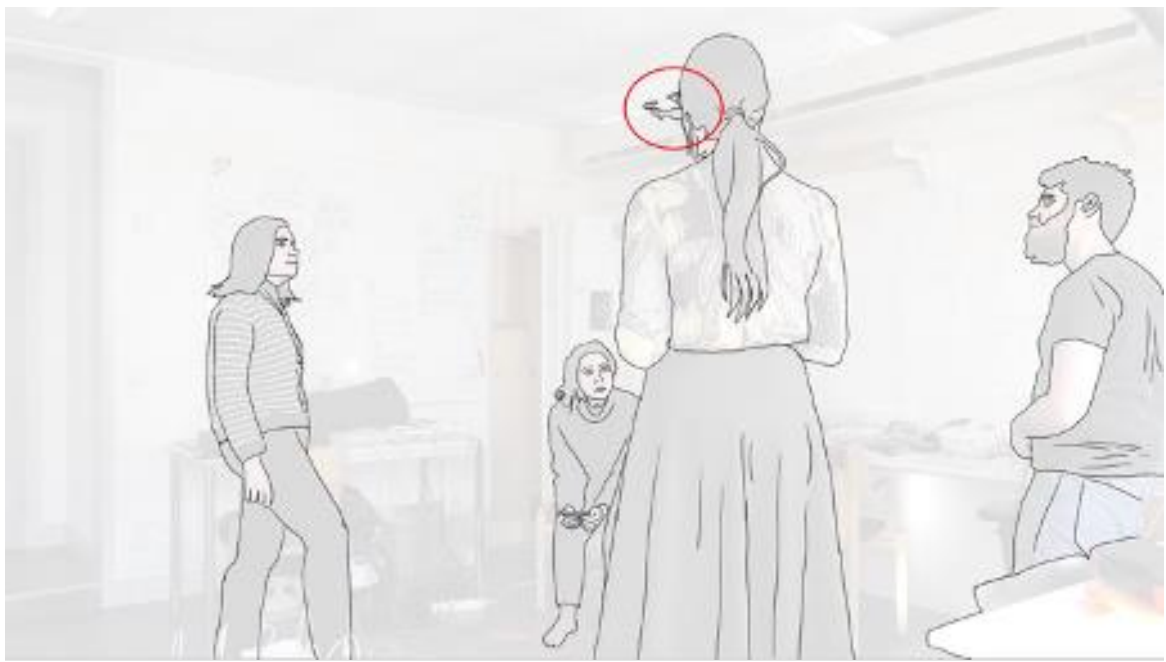

**Figure 8.** Maria (left) uses her foot to exert pressure on an inflatable cushion [out of shot] whilst Patrick (right) actuates the inflatable. Olivia (centre front) pilots the drone [circled in red]. Image reprinted from [46]

---

[1] The names of participants have been anonymised

In Figure 8, Maria uses her foot to exert pressure on an inflatable cushion attached to the floor (mimicking a foot pedal) whilst Patrick uses a pneumatic mechanism [64] to actuate the inflatable device in time with Maria's movements. Olivia pilots a small Tello EDU drone[2] using an application on her mobile phone.

During the exploration, it is mentioned that gaze may be an evocative design experience as the blinking red LED on the back of the drone appears similar to an eye. Following this observation, Olivia lowers the drone to be closer to eye-level with Maria and rotates the drone so that the blinking LED is facing Maria. Maria continues to enact how she imagines controlling the drone using the "foot pedal," exerting pressure on the inflatable in a rhythm similar to the sound made by the drone. Olivia begins to move the drone up and down in time with Maria's movements. At this point, Olivia notices that the drone is drifting uncomfortably close to Maria. She is concerned that it might crash into Maria, and she then attempts to direct the drone away from Maria using the controller on her mobile phone. However, as Olivia previously rotated the drone to face Maria, the orientation of the drone has now changed in relation to Olivia's relative position. When Olivia attempts to direct the drone away from Maria by dragging her fingers backwards across the controller, the drone responds by moving forwards towards Maria. Olivia immediately realises her mistake and tries to prevent the collision by redirecting the drone using the controller. In the same moment, Maria realises the drone is coming close to her, but does not back away from the drone as she believes that Olivia has control of it and that it would appear silly to express fear of the drone coming close to her. It is not until the collision is imminent that Maria realises that the drone is going to collide with her. There is insufficient time either for Olivia to redirect the drone or for Maria to move out of the collision path. The drone then collides with Maria, bumping into her chin before flipping up and becoming entangled in her hair and finally, shutting down. Olivia rushes to Maria and, apologising repeatedly, explains that she had not intended to direct the drone so close to Maria. Maria points to her chin where the impact caused a sharp pain but reassures Olivia that she is otherwise okay. Olivia disentangles the drone by dismantling the propellors and removing the body of the drone from Maria's hair. However, one propellor is firmly tangled in a knot of hair. Maria makes a light-hearted remark about the propellor in her hair which prompts laughter whilst Olivia works to untangle the final propellor. Once everyone is reassured that Maria is not seriously hurt or upset, embarrassed laughter and joking breaks out among the group.

**6.2 Applying Collision Trajectories to the Drone Collision**

In Figure 9, we sketch the interleaved trajectories that characterise this *mutually accidental* collision. As this was an undesirable or unwanted collision, here we show how our transitions can be applied to reveal the moments leading up to the breakdown between drone and operator, the reactions and responses to the collision, and the recovery from the collision.

---

[2] https://www.ryzerobotics.com/tello-edu

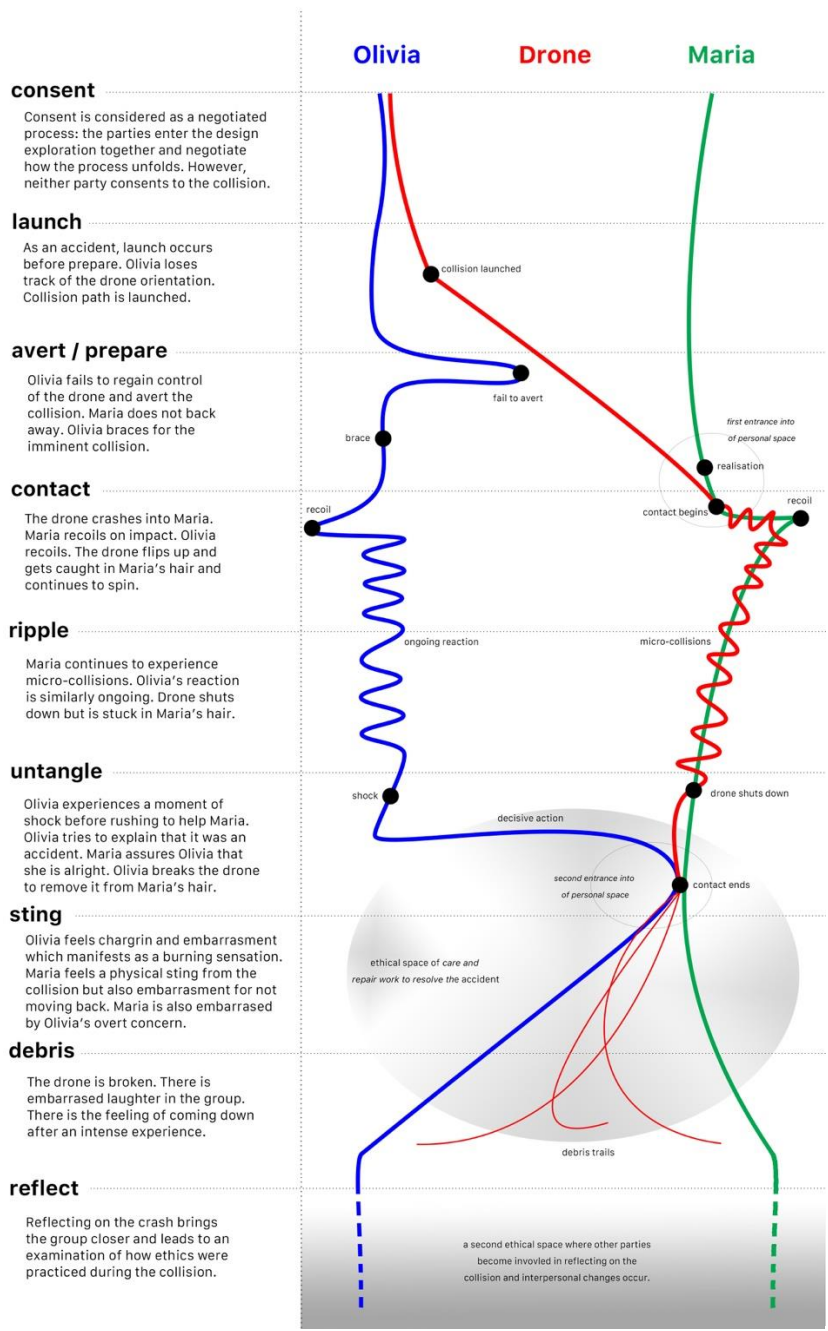

**Figure 9.** Analysing the trajectories involved in the drone collision

### 6.3 Reflections on the Drone Collision Trajectory

Our application of the collision trajectory reveals the complexity of this collision. Several important facets to the collision become evident when considered in this fashion. First, though Olivia and the drone enter the design exploration together, they begin to drift apart before the collision path is launched. This prompts us to focus on the moments leading up Olivia's losing orientation of the drone and which design features contributed to this divergence of trajectories. Second, there are multiple periods of so-called "non-action", for example; when Maria realises the collision is imminent; when Olivia braces for the collision; and when Olivia experiences a moment of shock before taking action to help Maria. Though these moments may be very brief; they were experienced as much longer by the collision parties which may have an impact on how parties react and respond to unfolding collisions. Third, even though Olivia is not directly involved in the impact, there is a considerable amount of reaction on her side of the trajectory. She, as the party responsible for the collision, also took action to resolve the situation. This reveals that teleoperators are far from being passive parties in these types of collisions. This prompts us to reflect on how autonomous robots, without an operator immediately present, would be able to effectively react and repair such events. Fourth, this is very evidently not a 'clean' collision, for example, after the initial impact Maria continues to experience micro-collisions as the propellors continue to spin. The "danger zone" of the collision is therefore not limited to the impact alone but stretches for a considerable stretch of the collision until Olivia fully removes the drone from her hair. Fifth, there are two moments where another party enters Maria's personal space; the initial impact of the drone and when Olivia rushes to help her. This reveals to us that the impact is not the sole factor when consideration violations of personal space, and attentiveness also should also be given to not "cause any further harm" by rushing blindly to resolve the situation. Debris trails were not caused by the impact itself, but rather Olivia's breaking of the drone when she removed it from Maria's hair. This shows us that the "fallout" from a collision should be considered as generated beyond the contact itself. Finally, reflection involved both conversations between Olivia and Maria to repair the emotional sting left by the collision, ultimately drawing the group close together, and eventually a detailed micro-phenomenological analysis as part of writing up research papers.

### 7 COLLISION CASE STUDY: CAT ROYALE

Here we present our second case study; that of a collision that occurred between a cat and a robotic arm that took place in the context of an interactive art installation. This second case study differs from the drone collision in several important respects. First, collisions were not accidental but rather occurred as a purposeful expression of play between the cat and the robotic arm. Second, unlike the drone collision, we have little access to the first-person experiences of the collision parties (except for the robotic operator). This analysis therefore relies more heavily on a third person view of the collision. Third, whereas the parties to the drone collision accidentally created a situation where a collision was possible, the artists who created Cat Royale intentionally fabricated a situation where a collision was likely, and even desirable, i.e., an environment where robotic arm engaged the cats in play but needed to also be safe. Another result of this is that the audience to Cat Royale were expecting to witness a collision or at least knew that it was possible. Finally, Cat Royale was extensively mediated by welfare officers and ethical overseers, whereas the drone collision occurred in a context negotiated between the parties involved. Our purpose in presenting two starkly different collisions, is to show the utility and flexibility of the collision trajectory.

#### 7.1 Description of Cat Royale

Cat Royale is an artwork created by the artists Blast Theory to explore the question of trust in autonomous systems [57]. Specifically, it invited an audience to consider the conditions under which they would trust a robot to play with cats. Three cats spent six hours a day for twelve days in a purpose-built enclosure that, following guidance from feline welfare experts, was designed to be, in the artists' words, a 'cat utopia' providing an ideal environment for them. The enclosure included multiple secluded dens for sleeping, feeding stations, a water fountain, litter trays and high walkways and perches from which they could view it. At its centre was a robot arm that, once every ten minutes, attempted to enrich the cats' lives by playing with them. The robot was a Kinova Gen3 Lite, an ultra-lightweight small robot arm equipped with a non-interchangeable two-finger gripper, chosen due to its inherent safety-by-design through low speed (25 cm/s) and payload (0.5 kg) limits to work near humans (and in this case cats too). Playing with the cats involved picking up a variety of toys from four nearby racks before moving them in ways that would attract the cats' attention and ultimately lead to their physical engagement. The robot arm could only move relatively slowly compared to the reaction times and movements of the cats and was securely mounted to the floor so that it could not be moved (important for its calibration to be able to pick up and replace the toys on the racks) or fall over. Human operators continually monitored the robot from a control room through a one-way mirror. They had the capability to pause it at any time, by releasing a deadman's switch (a button that had to be constantly held down for the robot to move); switch over to alternative pre-programmed movement sequences; or try to improvise new movements if necessary.

Over the course of the twelve days the robot experimented with over 500 games and treats. Many of the toys such as the feather bird shown in Figure 10 (lower left corner) were attached to a rod via an elastic string so that they could be bounced and jigged in an attractive manner and to enable the cats to physically bat, claw and grasp them. Human observers scored the cats' engagement with each game using the feline Participation in Play Scale [15], feeding the results into a decision engine that slowly learned the cats' individual preferences. The whole experience was filmed using eight cameras positioned within the environment The resulting material was then edited to create an eight-hour long film to subsequently be shown as a public installation in galleries. The project received extensive advice from veterinarians, experts in feline behaviour and animal welfare organisations as well as ethical approval from the University's Animal Welfare and Research Ethics Board, Veterinary Review Board and Computer Science Review Board, following an extensive ethical review process as discussed in [3]. A Cat Welfare Officer continually monitored the cats for signs of distress or stress and could require the cats to be withdrawn if necessary, logging their observations every 15 minutes. The cats' owner was also present on site. A subsequent study of Cat Royale reflected on how designing the cat's interactions with the robot needed to take account of a wider 'multispecies world' that included both the careful design of the surrounding enclosure alongside the roles and procedures for humans to monitor events and wrangle to robot [56].

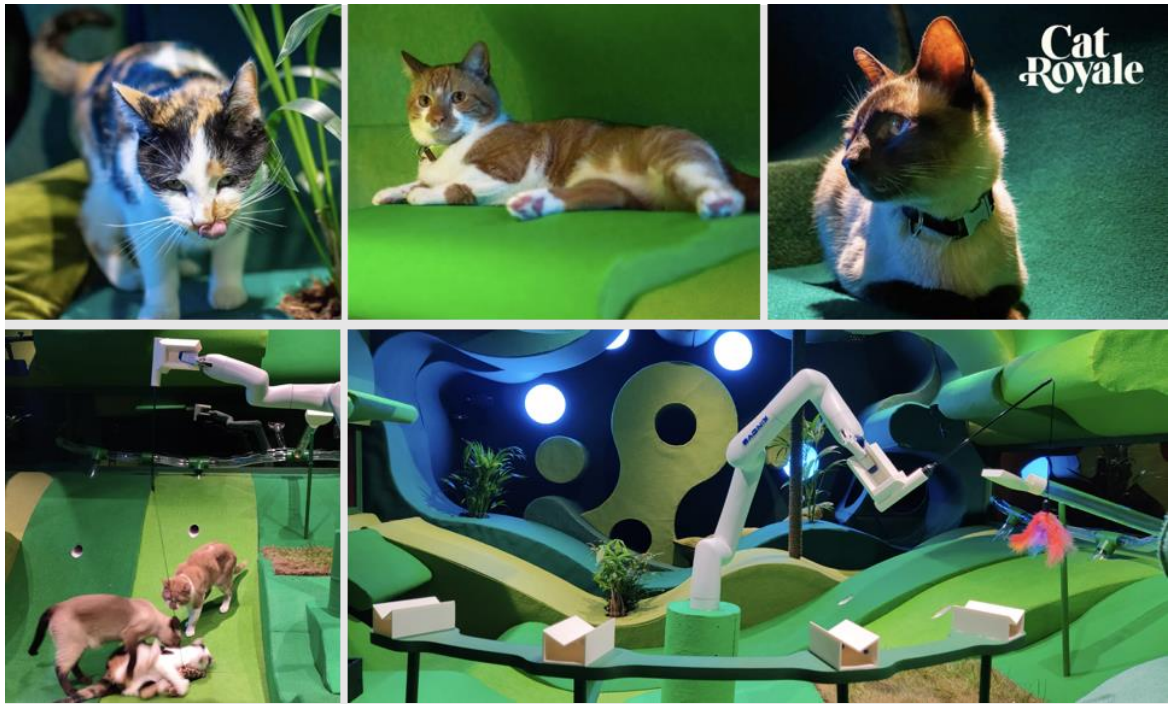

**Figure 10.** Images from Cat Royale. Top: the three feline stars of Cat Royale, Clover, Pumpkin and Ghostbuster. Bottom: playing with the robot and the robot offering the feather toy.

We now focus on one notable play sequence that represents many of the interactions we witnessed in microcosm. It occurred relatively late on (day ten) and involved Clover (one of the cats) physically battling for possession of the feather toy for several minutes. By this time, Clover had become highly interested in, and increasingly assertive with, the robot, especially when it offered the feather bird toy. The play sequence begins with Clover approaching the robot, following it as it drags the toy across the floor, and assuming a pouncing position as shown in Figure 11 (a). She then physically engages with the feather toy by batting it with her paws (b) and grabbing it in her jaws (c). A series of tugs-of-war ensue in which she repeatedly grabs the toy and pulls strongly against the robot (c). These visibly stress the robot arm which can be seen bending against her force to the point where the operators decide the robot should let go of the toy (d). The problem is when and how to do this safely so that the toy doesn't fly back and hit Clover. Eventually they find a suitable moment when Clover has chosen to release her grip and trigger the robot to drop the toy (e). Clover now picks up the toy in her mouth (f) and with some difficulty drags it around the robot (g), over a lip in the floor (h), and past the water fountain, which it pulls over (i), as she carries it away from the robot.

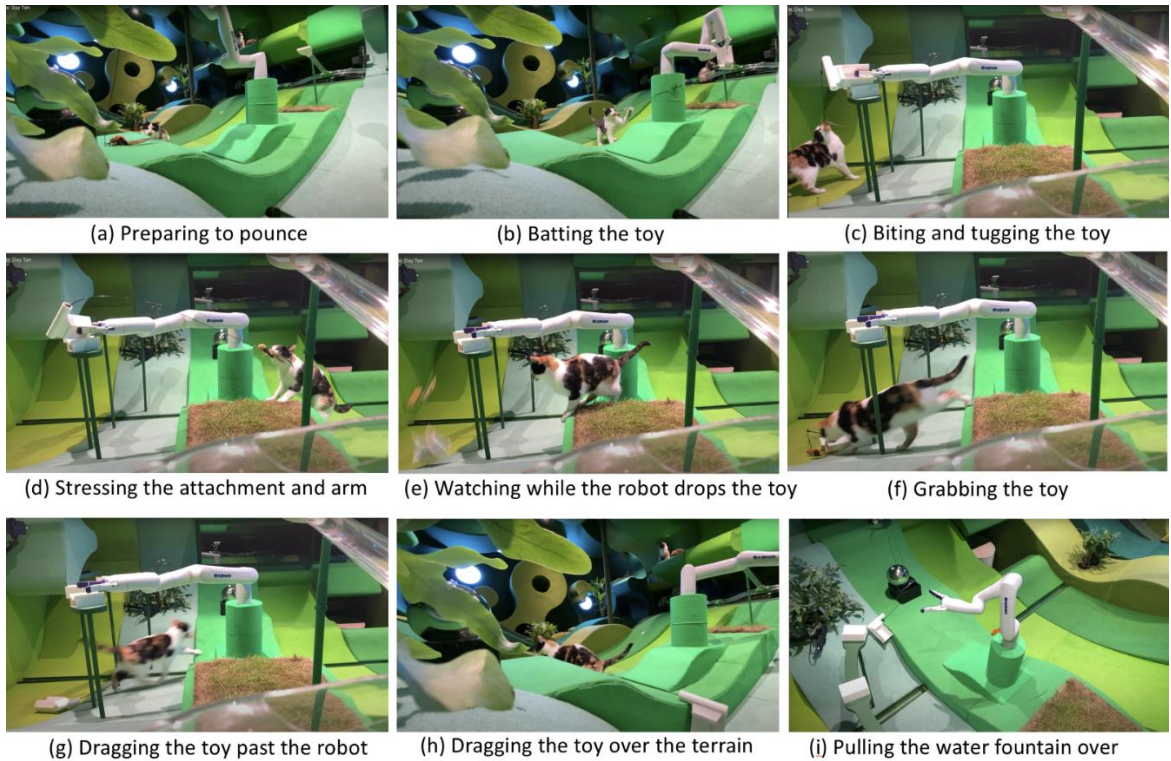

**Figure 11.** An extended collision in which Clover wrestles the bird toy away from the robot

### 7.2 Applying Collision Trajectories to Cat Royale

In Figure 12, we visualise the trajectories which characterise this *mutually managed* collision. As this was a desirable or intentional collision, here we show how our transitions can be applied to reveal how the designers of Cat Royale orchestrated a playful interaction between cat and robot whilst maintaining a high level of safety, as well as the role of the operator in facilitating a safe beginning and end to the collision experience.

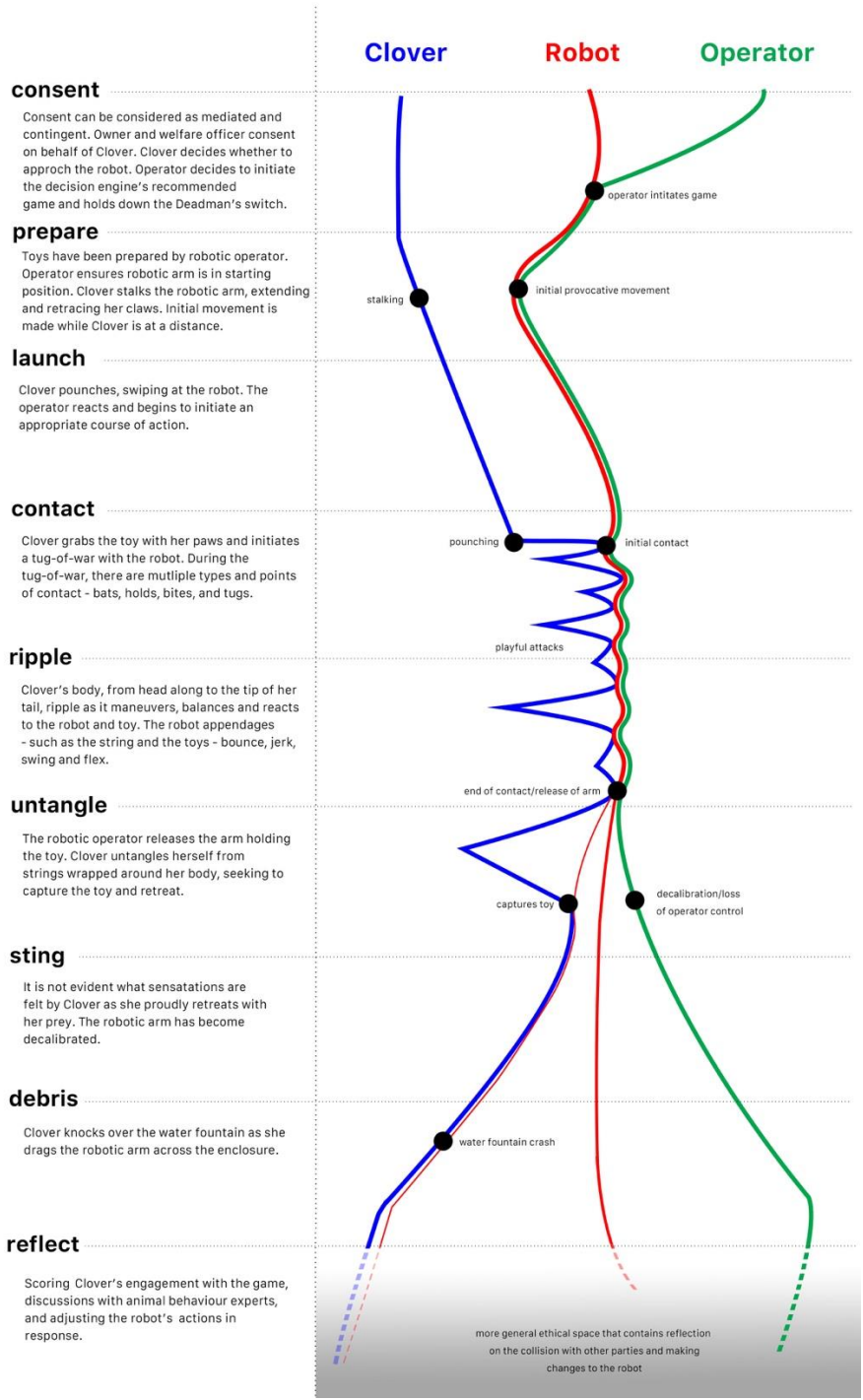

**Figure 12.** Analysing the trajectories involved in Clover's game with the Feather Bird Toy

### 7.3 Reflections on Cat Royale Trajectory

Our application of the collision trajectory reveals several important facets on this collision. First, the trajectory clearly depicts how Cat Royale's designers orchestrated a mutually managed collision where the experience gradually unfolded for both parties. This began with a carefully negotiated back and forth between Clover and the robot operator to negotiate initial consent and preparation, with the robot initially signalling its intent by picking up the toy, and Clover hers by dropping from her high perch to approach the robot in a crouch. Consent continued to be negotiated throughout and at one point became challenging as Clover took control and placed the robot operator in a situation in which they wanted to disengage but were unable to. Second, we can see from the different trajectories how Clover, the robot and the operator all have different capacities to act in the unfolding collision. It takes time to change the course of the robot's movement once started due to both reaction times, but also due to the time it takes the operator to react, formulate and initiate an alternate suitable course of action. Given that the cats are far more agile and have far sharper reactions than the robot, many of its actions have an essentially ballistic quality when it is close to the cats in the sense that it cannot control the physical contacts that may ensue (whereas the cats usually can, e.g., by pulling back or ducking out of the way). This is shown in the sharpness of the trajectories and the speed at which they change direction as well as how the operator trajectory is dragged along with Clover's attacks on the robot until they find an opportune moment to disengage. Attachments also get wrapped around the robot arm and nearby stands at which point the operator may have to spend considerable time and effort working out a series of moves that will disentangle it. Third, this trajectory illustrates how debris potentially make it harder for parties to execute its moves if unanticipated obstacles were now in the way, such as Clover's secondary collision with the water fountain. Finally, there were multiple points of reflection in Cat Royale: each game was immediately rated by the artists to train the decision engine to recommend further games; new games were devised and robot movements were authored in response; animal welfare experts monitored the cats' behaviours, feeding back into a daily discussion involving a vet; and ultimately analysis of various collisions were written up in research papers.

These two contrasting case studies reinforce the idea of extended collision trajectories (lasting several minutes) in which the experience of collision begins long before any physical contact and significantly extends beyond first contact. The drone trajectory reveals significant insight into the recovery process of an accidental collision, while Cat Royale provides an example of designing to balance desirable (enabling pleasurable play) and undesirable (avoiding injury to cat or robot) aspects of collision.

## 8 TANGLES (ENTANGLED TRAJECTORIES)

Our analyses of the Drone Collision and Cat Royale reveal the complexities of 'real world' collisions. Such collisions may draw in multiple parties, involve repeated contacts as bodies bounce against each other, demand the ongoing negotiation of consent and control throughout, deliver significant emotional as well as physical experiences, and require significant untangling. While we found that the simple soma collision trajectory with its nine transitions from Section 5 provided a useful foundation for expressing important aspects of such collisions, it felt insufficient to capture their inherent messiness. In short, real-world collisions do not unfold in such a straightforward linear way but rather involve deep and ongoing entanglements between multiple parties, leading us to consider the idea of entangled trajectories, or tangles for short.

*Tangles* considers collisions between parties in terms of the multiple ways in which they can become *entangled* and *disentangled*. We see this not only in terms of physical contact (momentary contact, multiple blows, extended impact), but also in terms of emotion (a shared moment of shock, an argument about blame) and relationships between parties (breakdown following an accident, bonding after a shared experience). As a collision unfolds, parties become tangled up—literally and figuratively—bound together in ways that give rise to the possibility of further collisions or different emotional responses. In other ways, they simultaneously become disentangled; objects break apart such as when a robot lets go of a toy; a human operator becomes separated from a drone as they lose control; or trust may be lost between parties. Moreover, other people and objects may also subsequently become entangled in the collision. Spectators or other third parties may become involved in the unfolding experience as may other objects, potentially including other robots. The result is that collisions give rise to *re-entanglements* as multiple parties become tangled up in new configurations, potentially with the key transitions we noted earlier recurring. In short, it is more realistic, if more challenging, to envisage collision experiences as *tangles* rather than simple linear trajectories we initially envisaged in Figure 6.

A tangle can therefore be defined as *a messy interleaving of multiple parties' soma trajectories as they become drawn into a collision that results in various physical and emotional re-entanglements among them*. As a visual summary, Figure 13 offers a 'zoomed out' sketch of such a *tangle* between a robot, local human, remote operator who become both disentangled and re-entangled throughout.

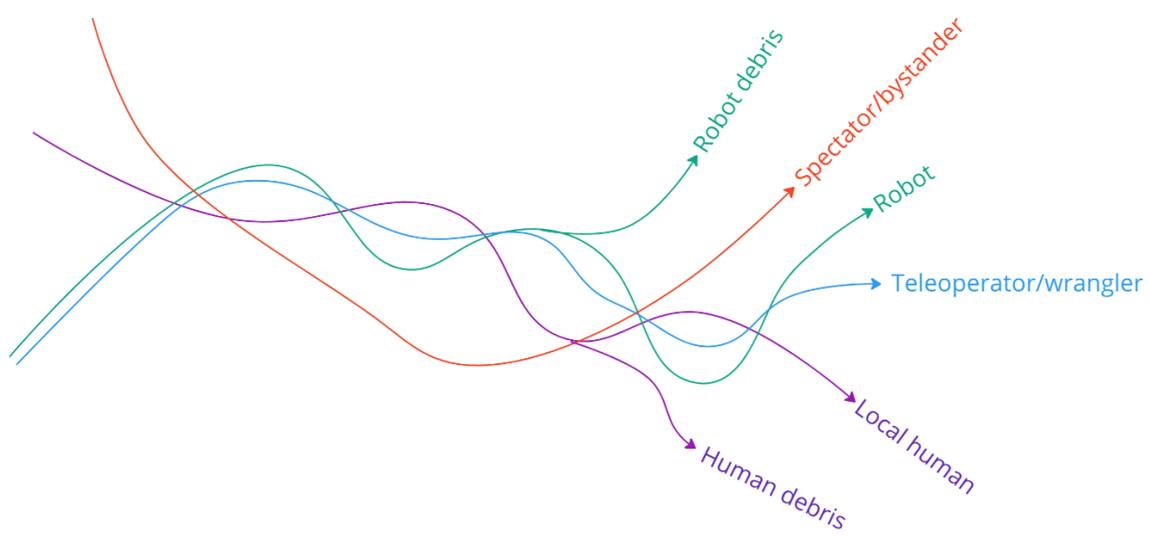

**Figure 13.** Collisions as messy tangles between parties

We now propose three ways in which this formative notion of tangles might be useful: 1) as a lens for analysing previous collisions; 2) informing strategies for designing future ones; and 3) broaching the ethical complexities of collisions.

**8.1  Tangles as an analytical lens**

As collisions between humans and robots have so far been approached as a risk to be actively minimised and avoided, there are few tools in HCI available to dissect the experience that might result from a collision. We therefore intend for tangles to provide an analytic lens through which researchers can better understand collisions. Tangles offer a practical set of transitions as a starting point to break down a collision and analyse how it unfolded, including the circumstances leading up to a collision and reverberations that follow. Tangles calls attention to the physical, emotional and relational dimensions of collision. Finally, it encourages researchers to consider other parties beyond those who directly come into contact as also being involved. As we have seen in this paper, these provide themes and sub-themes for analysing a variety of collision data, including observations and first-person accounts, both for the breakdowns involved in *unwanted or accidental* collisions, or for the enablers of *intentional or desirable* collisions. We particularly encourage researchers to consider sketching soma trajectories as we have done with the above examples as even rough sketches call attention to the tangled nature of collisions.

## 8.2 Tangles as a design approach

We also see potential for tangles to be employed as a generative design tool. Though this has not yet been the primary focus of our work, we offer two novel design strategies for generating collision experiences based on our approach of tangles.

Building on our observation in the previous section, our first strategy is *to exaggerate the emotional anticipation and consequences of collision.* One might deliver powerful yet safe collision experiences by trading off actual physical contact for heightened anticipation and emotional consequences. For desired collisions, this might mirror the design of the thrilling, but uncomfortable, interactions found in rollercoasters and similar rides in which people relish fear beforehand and are given the means to remember and tell stories afterwards, but where the actual physical experience is carefully constrained. The same strategy might also apply to undesired collisions, where it may be better to anticipate them if possible and perhaps to exaggerate minor contact and near misses afterwards to help people learn about danger and mistakes.

Our second strategy, which we suggest is somewhat counter to conventional thinking in robot design, is to deliberately *extend robot bodies to become more tangly and hence more collidable,* a strategy previously explored in [29]. Our initial workshop drew our attention to the idea that collisions might be experienced as skin, muscle or bone, which led to the approach of using various attachments to mediate and extend collisions with robots including balloons, vibrators, and Spot's whippy tail. This approach was evident in Cat Royale's deployment of many different toys, on rods and strings, and as projectiles (balls). Such attachments make robots more "collidable" in the dual sense that collisions become more likely but also become softer. They also mirror previous research into how people dress robots with skins, clothes, costumes and bling [19]. Robot bodies might also be augmented to deliver ripples, stings and debris. They might make exaggerated rippling movements or perhaps glow or throb to show that collisions have happened. They might be designed to easily break and throw of debris as a way of absorbing impact (as cars and other vehicles do), to again reveal that collisions have happened so that people can learn about them and account for them.

## 8.3 Tangles as a lens to broach ethical complexity

As discussed in previous work [46, 39], a growing body of research suggests that collisions are a far more ethically complex phenomenon than is generally considered by interaction designers, wrapped up with, not only concerns of risk, safety and accountability, but also the freedom to learn and play, inhabit different spaces, and participate freely in many sports and other enjoyable activities. Our purpose here is not to argue that enabling or encouraging collision between humans and robots is always desirable or ethically justified, but rather to call for a deeper consideration of a complex issue that is often simply treated as something to totally avoid⎯especially as total avoidance may not always be possible or desirable. Here, tangles serve to cast a more nuanced ethical lens on the unfolding of a collision as well as consideration of the parties involved. Our analysis of both Cat Royale and the Drone collision reveals complex and negotiated processes of consent that temporally unfold in relation to the changing constellations of those entangled in the collision experience. Cat Royale involved both mediated and contingent consent as discussed in [36]. Mediated consent was given by suitably qualified humans (the owner and cat welfare specialist), whereas contingent consent was judged to be given by the cats themselves according to whether and how they chose to approach the robot. Both were negotiated on an ongoing basis as interactions unfolded. For the drone collisions, this ongoing negotiation of consent was apparent in how Olivia and Maria together managed the unfolding design exploration that led to the collision and when that consent is withdrawn. Further, Olivia and Maria both reflect that the collision—though unwanted and accidental—was an event that prompted care and repair, led to their relationship becoming closer, and became an ultimately important experience for them both. By highlighting Olivia and Maria's *re-entanglement*, tangles may help unpack how ethically tricky situations (in this case, the cause of unintentional harm) are successfully resolved and help envision how and when robots might respond in such situations. Further, by engaging in and unpacking the experience of a collision that both parties considered ethical, such as the orchestrated collisions in the workshop, we can cultivate a better sensibility towards the context and qualities that are critical to designing for collisions in an ethical manner and how/when people are empowered to withdraw [20].

Ethically speaking, tangles prompt us to *reconsider collisions to accommodate a wide ecology of roles*. tangles show us how collisions extend beyond what/who might normally be considered as the colliding parties, i.e., the robots and humans who come into direct physical contact. In the case of teleoperated robots such as the Double 3, the Drone, Spot, and to some extent Cat Royale's robot arm, collisions also entangle their remote operators (or temporarily disentangle the operators from the robots if they lose control). Even with robots that are not overtly teleoperated, there are likely to be local wranglers or remote overseers with responsibility for monitoring them and potentially intervening in challenging situations: for example: care robots will work alongside care assistants and inspection robots will be locally or remotely monitored. Finally, there are those who witness collisions, where we can distinguish knowing audiences who have some understanding of what is happening (the viewers in Cat Royale, or perhaps other people who are in a group with a telepresence robot), from what are termed 'unwitting bystanders' [58] who happen to be in the vicinity but are outside of the so-called performance frame [2] and so may have little sense of what is going on.

## 9. CONCLUSION

We have argued that, as robots and similarly physically actuated technologies migrate from specialist, tightly controlled environments into everyday life, we need to fundamentally reappraise what it means to collide with them. In everyday use, collisions with robots will become inevitable, mundane, and more of a nuisance than an extreme event. They might also become playful, joyous and expressive, enhancing the aesthetic experience of being an embodied human. Here, we have reappraised collision from a somaesthetic perspective, employing a soma design approach to explore the mind-body experience of colliding with various kinds of robotic devices. This yielded insights about what it feels like to collide and how to improvise new collision experiences. Reflections on these early explorations gave rise to two initial concepts that helped sensitise us to the nuances of collision experiences: 1) that an extended collision experience can be considered to be interleaving of two parties' soma trajectories and 2) that these trajectories pass through nine key transitions in anticipation of and consequential to any point of contact: consent, preparation, launch, ripples, stings, untangling, leaving debris, and reflecting. We then employed these initial concepts to analyse two examples of rich real-world collision experiences: an accidental collision between a human and a drone, and an intended collision between a cat and a robot arm that played with her. Our analyses led us to extend the initial idea of a collision trajectory into the idea of *tangles* that captures the messiness of ongoing entanglements between multiple parties to a collision. We propose that this formative notion of tangles might provide a sensitising concept to guide the post-hoc analysis of collision experiences, suggest novel design strategies for future ones (exaggerating emotional aspects and extending bodies with attachments), and can help negotiate the ethical complexities involved in collision. Future work might seek to validate and refine the framework by applying it to further post-hoc analyses of collisions and/or the design of new technologies to enable more collidable robots.

## ACKNOWLEDGEMENTS

We are grateful to the following bodies for supporting this research: The Engineering and Physical Sciences Research Council (EPSRC) through Grants EP/Z534808/1 (Turing AI Fellowship on Somabotics - Creatively Embodying Artificial Intelligence) and EP/S023305/1 (EPSRC Centre for Doctoral Training in Horizon: Creating our Lives in Data); the Marianne and Marcus Wallenberg Foundation (WASP-HS grant MMW 2019.0228); the Strategic Research Foundation (SSF) of Sweden (grant CHI19-0034 Hardware for Energy Efficient Bodynets); and Centrum for Idrottsforskning (grant 2021-04659).

## DATA ACCESS STATEMENT

The data that support the findings of this study may be available on request from the corresponding author, [Steve Benford]. The data are not publicly available due to their containing information that could compromise the privacy of research participants.

## STATEMENT OF PREVIOUS RESEARCH

The approach of designing to embrace collisions (rather than only avoiding or mitigating them) was previously reported in a short alt.chi paper at CHI 2023 [39]. However, that paper articulated the challenge in terms of revisiting the familiar idea of a risk matrix, which is not included as a contribution in this paper, which instead focuses on the different approach of extending soma trajectories. The report from the soma design workshops in Section 4 of this paper and the conceptual explorations of soma collision trajectories and tangles in Sections 5 and 8 respectively have not been reported elsewhere in any form. The drone collision case study in Section 6 was previously reported in a paper at CHI 2022 [46], but what is presented here is a substantial reanalysis of that work based on the concept of soma trajectories that has not been previously published. A short overview of the Cat Royale cast study (Section 7) has previously been published in [51], while two papers were published CHI 2024, one focused on the ethical approval process [3] and a second on the holistic design of the 'robot world' [56] .None of these papers focuses on the matter of collision, and as with the Drone case study, the analysis here in terms of soma trajectories is completely novel.